\def\year{2022}\relax
\documentclass[letterpaper]{article} 
\usepackage{aaai22}  
\usepackage{times}  
\usepackage{helvet}  
\usepackage{courier}  
\usepackage[hyphens]{url}  
\usepackage{graphicx} 
\usepackage{natbib}  
\usepackage{caption} 
\DeclareCaptionStyle{ruled}{labelfont=normalfont,labelsep=colon,strut=off} 
\usepackage{enumitem}\setlist{nolistsep,leftmargin=1.5em}

\newcommand{\citeauthoryearp}[1]{\citeauthor{#1}~(\citeyear{#1})}

\usepackage{algorithm}
\usepackage{algorithmic}

\usepackage{newfloat}
\usepackage{listings}
\floatstyle{ruled}
\newfloat{listing}{tb}{lst}{}
\floatname{listing}{Listing}

\begin{document}

\title{tile2tile: Learning Game Filters for Platformer Style Transfer}

\author{
Anurag Sarkar and Seth Cooper
}

\affiliations{
Northeastern University\\
sarkar.an@northeastern.edu, se.cooper@northeastern.edu
}

\maketitle

\begin{abstract}
We present \textit{tile2tile}, an approach for style transfer between levels of tile-based platformer games. Our method involves training models that translate levels from a lower-resolution sketch representation based on tile affordances to the original tile representation for a given game. This enables these models, which we refer to as filters, to translate level sketches into the style of a specific game. Moreover, by converting a level of one game into sketch form and then translating the resulting sketch into the tiles of another game, we obtain a method of style transfer between two games. We use Markov random fields and autoencoders for learning the game filters and apply them to demonstrate style transfer between levels of Super Mario Bros, Kid Icarus, Mega Man and Metroid.
\end{abstract}


\newcommand{\XFIGUREtileaff}{
\begin{figure}[t!]
\centering
\setlength{\tabcolsep}{2pt}
\begin{tabular}{cccc}
\includegraphics[width=0.1\textwidth]{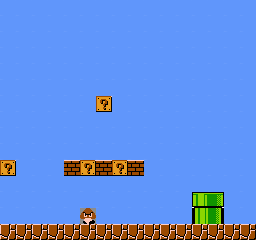}
&\includegraphics[width=0.1\textwidth]{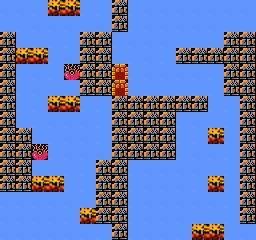}
&\includegraphics[width=0.1\textwidth]{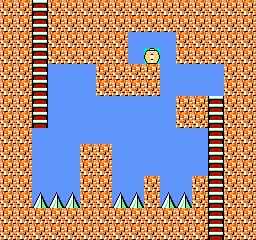}
&\includegraphics[width=0.1\textwidth]{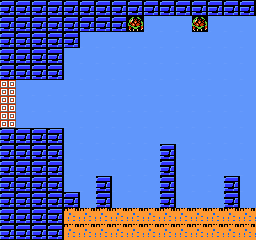}
\\
\includegraphics[width=0.1\textwidth]{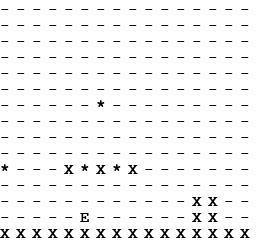}
&\includegraphics[width=0.1\textwidth]{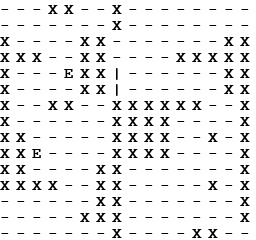}
&\includegraphics[width=0.1\textwidth]{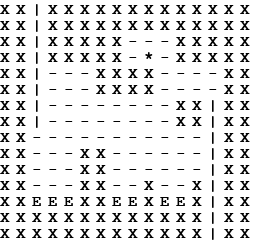}
&\includegraphics[width=0.1\textwidth]{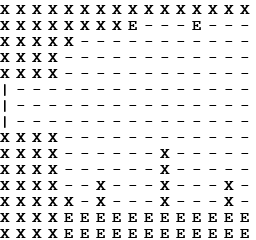}
\\
SMB & KI & MM & Met\\
\end{tabular}
\caption{\label{XFIGUREtileaff} Example segment from each game shown using original tiles and corresponding sketch representation.}
\end{figure}
}

\newcommand{\XFIGUREmrf}{
\begin{figure}[t!]
\centering
\includegraphics[width=0.3\textwidth]{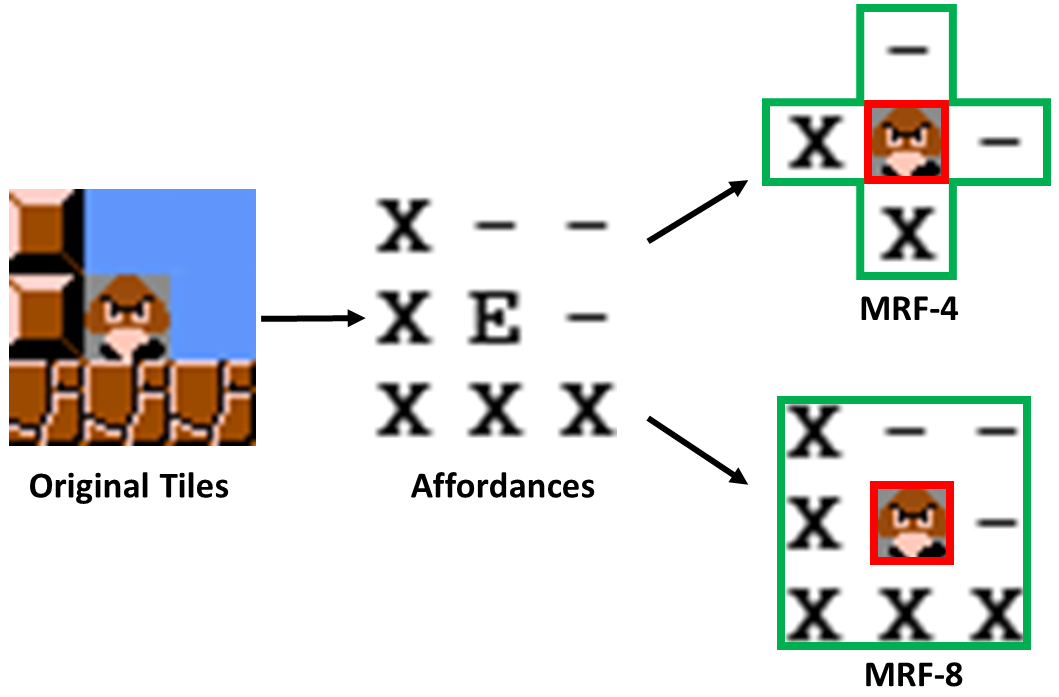}
\caption{\label{XFIGUREmrf} Markov random field training process. Green context sketch tiles were used to predict red game tile.}
\end{figure}
}

\newcommand{\XFIGUREmmtomet}{
\begin{figure}[t!]
\centering
\includegraphics[width=0.8\columnwidth]{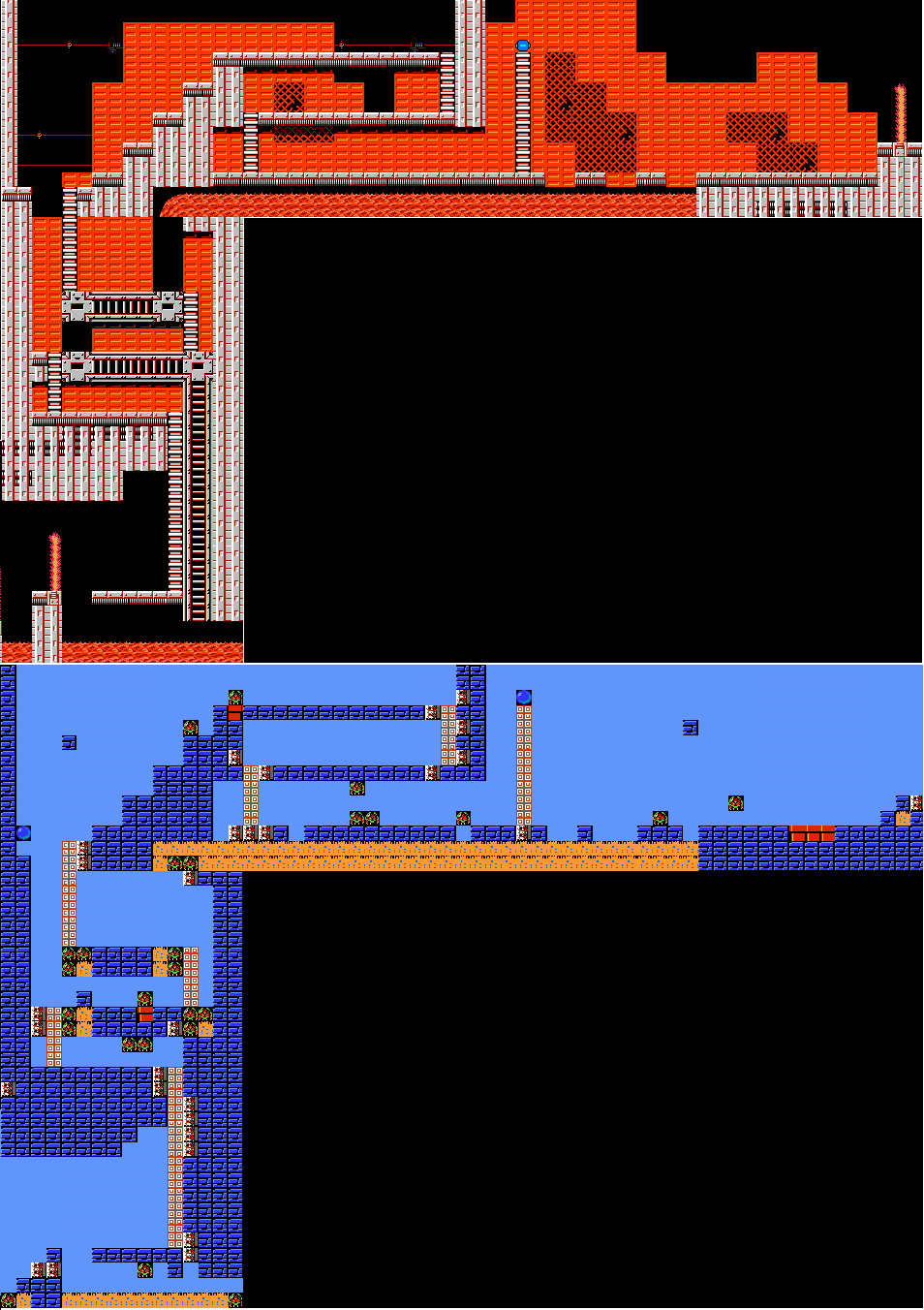}
\caption{\label{XFIGUREmmtomet} MM (top) to Metroid (below) using MRF-8}
\end{figure}
}

\newcommand{\XFIGUREsmbtoki}{
\begin{figure*}[t!]
\centering
\includegraphics[width=0.6\textwidth]{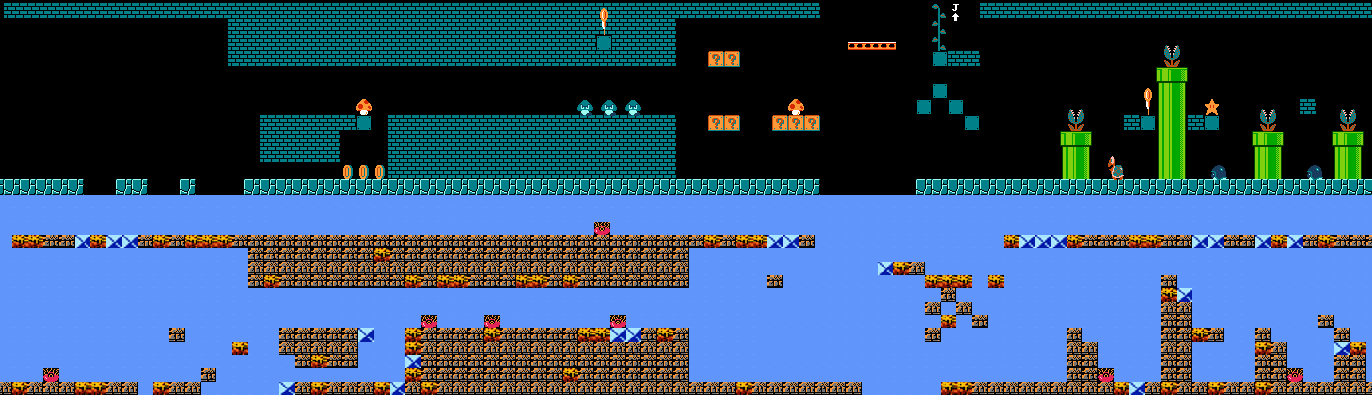}
\caption{\label{XFIGUREsmbtoki} SMB (top) to KI (bottom) using MRF-8}
\end{figure*}
}

\newcommand{\XFIGUREae}{
\begin{figure*}[t!]
\centering
\includegraphics[width=0.75\textwidth]{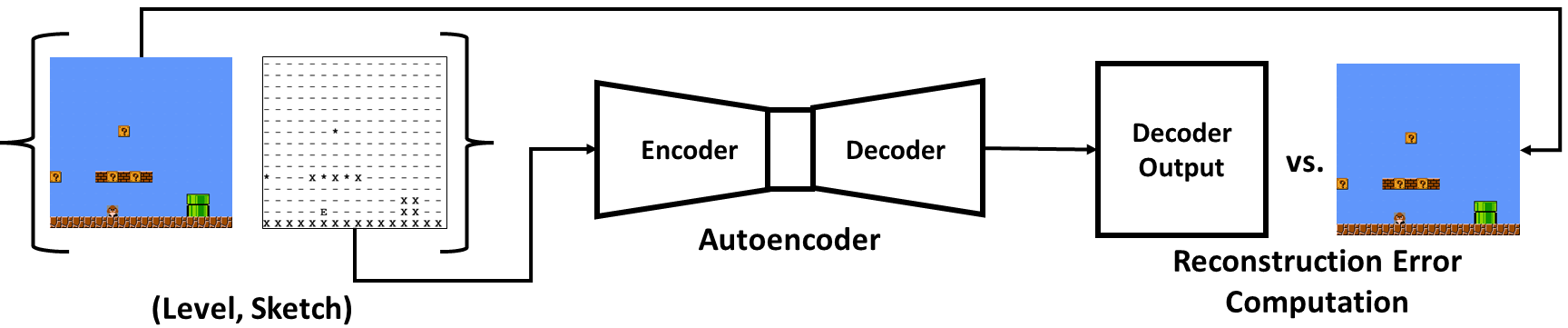}
\caption{\label{XFIGUREae} Autoencoder training process.}
\end{figure*}
}

\newcommand{\XFIGUREtiletile}{
\begin{figure*}[t!]
\centering
\includegraphics[width=0.8\textwidth]{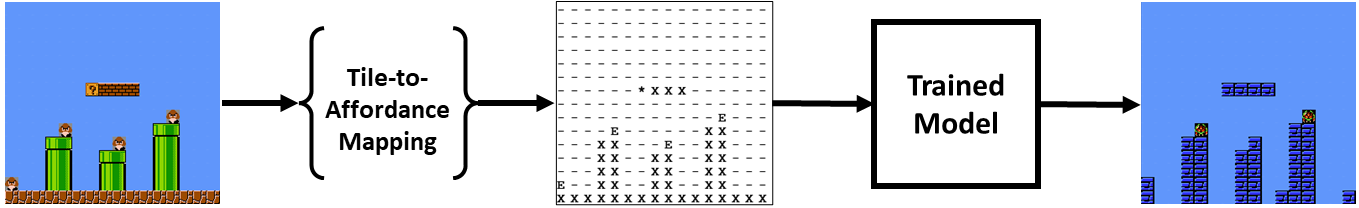}
\caption{\label{XFIGUREtiletile} tile2tile style transfer pipeline. In this example, an original Mario level is converted to sketch form using a tile-to-affordance mapping and then a trained model converts the sketch to a Metroid level.}
\end{figure*}
}

\newcommand{\XFIGUREtilehist}{
\begin{figure*}[t!]
\centering
\begin{tabular}{ccc}
\includegraphics[width=0.275\textwidth]{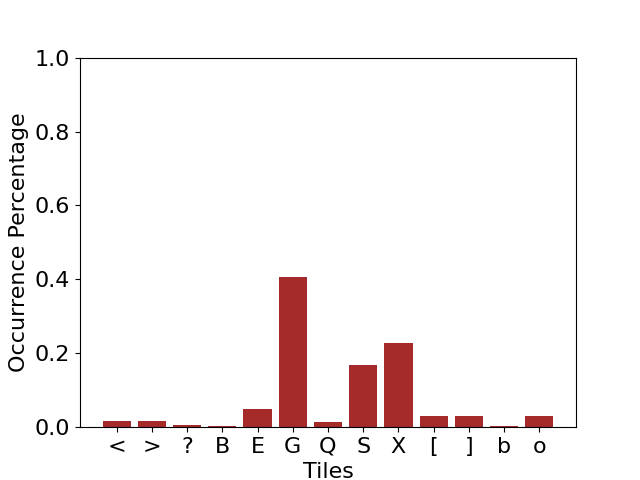}
&\includegraphics[width=0.27\textwidth]{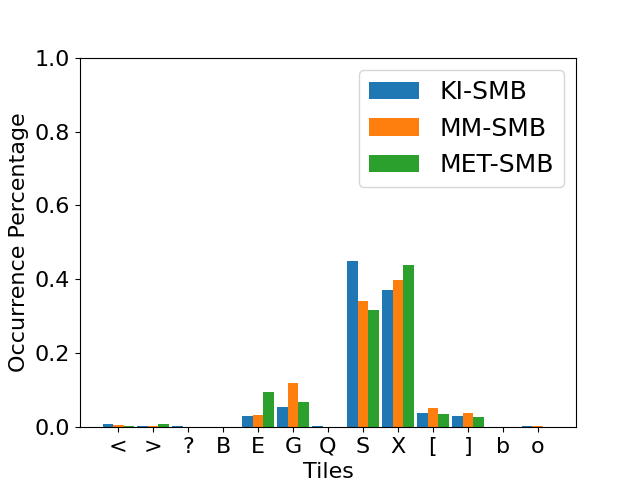}
&\includegraphics[width=0.27\textwidth]{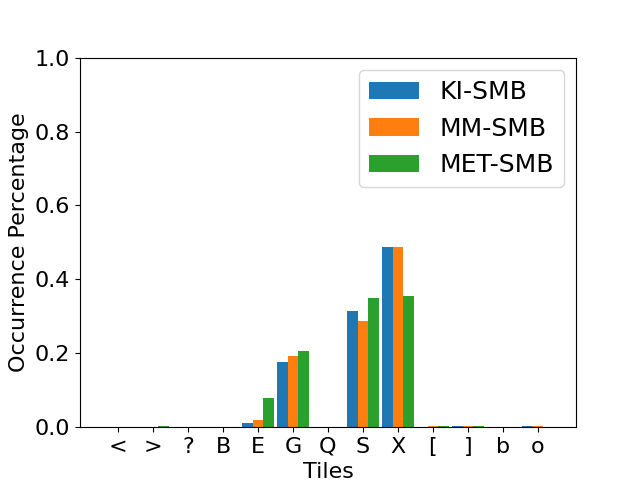}\\
Original SMB & \multicolumn{2}{c}{SMB as Target} \\
\includegraphics[width=0.27\textwidth]{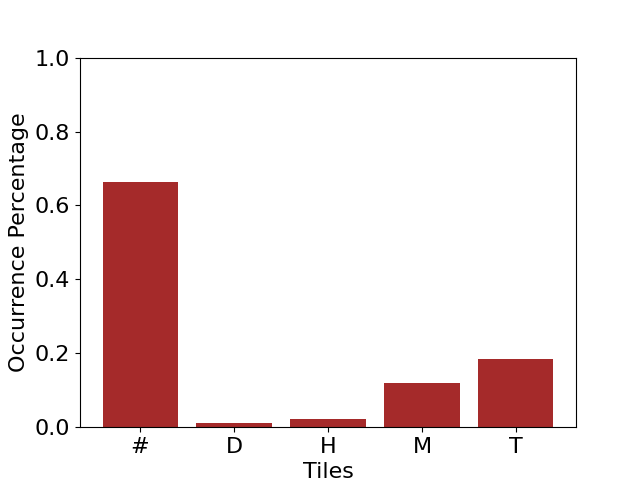}
&\includegraphics[width=0.27\textwidth]{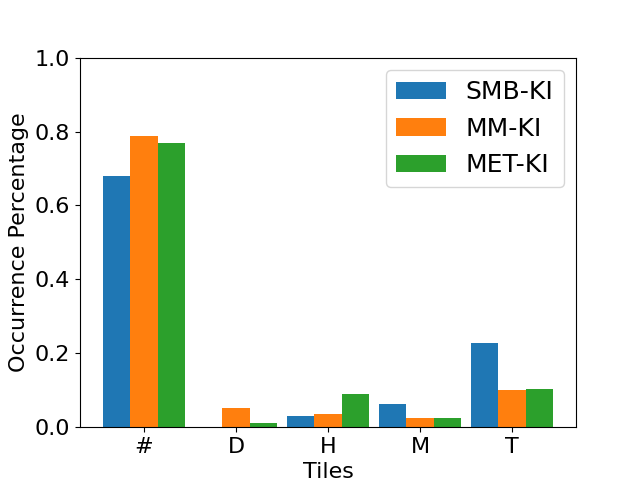}
&\includegraphics[width=0.27\textwidth]{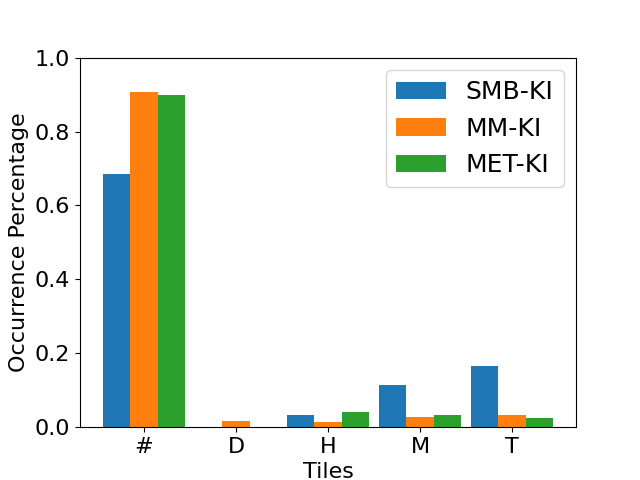}\\
Original KI & \multicolumn{2}{c}{KI as Target} \\
\includegraphics[width=0.27\textwidth]{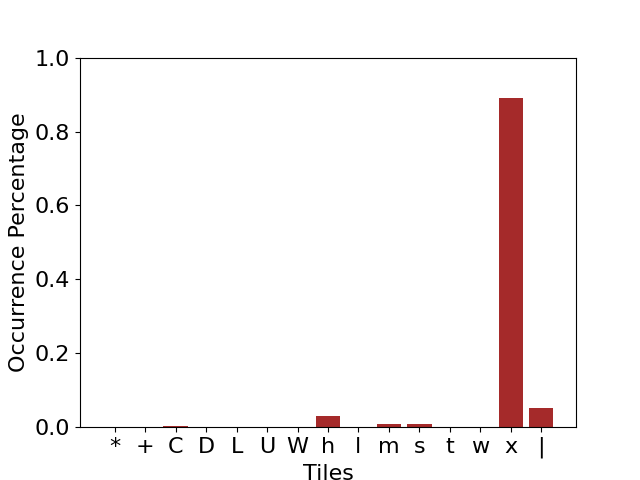}
&\includegraphics[width=0.27\textwidth]{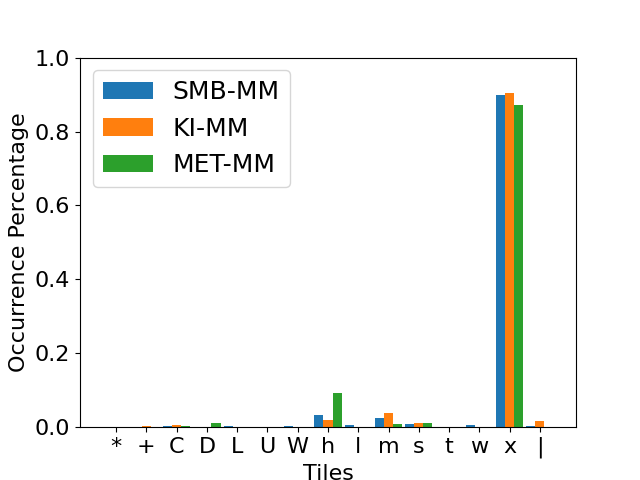}
&\includegraphics[width=0.27\textwidth]{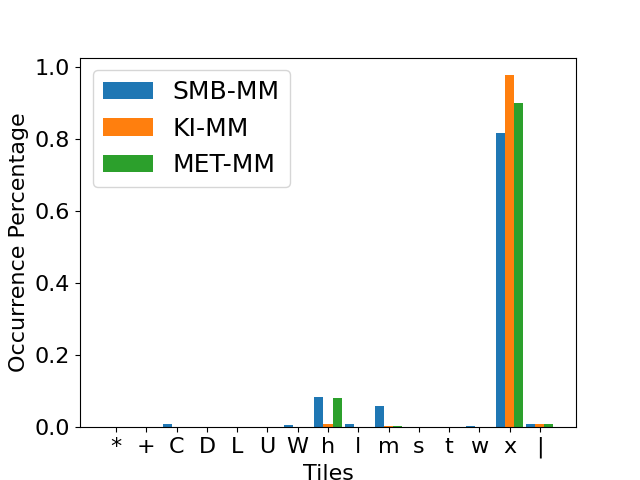}\\
Original MM & \multicolumn{2}{c}{MM as Target} \\
\includegraphics[width=0.27\textwidth]{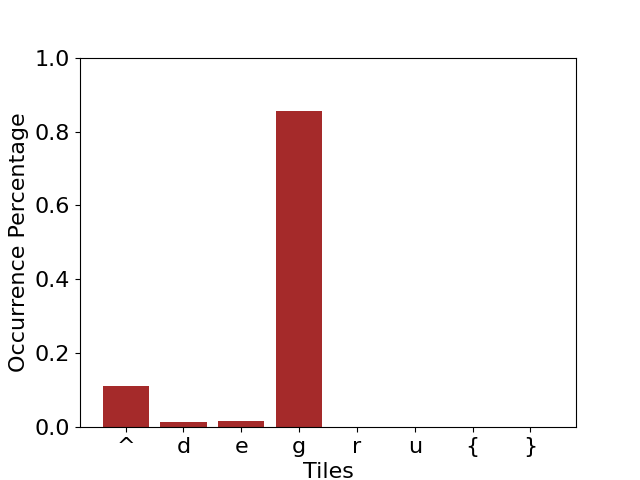}
&\includegraphics[width=0.27\textwidth]{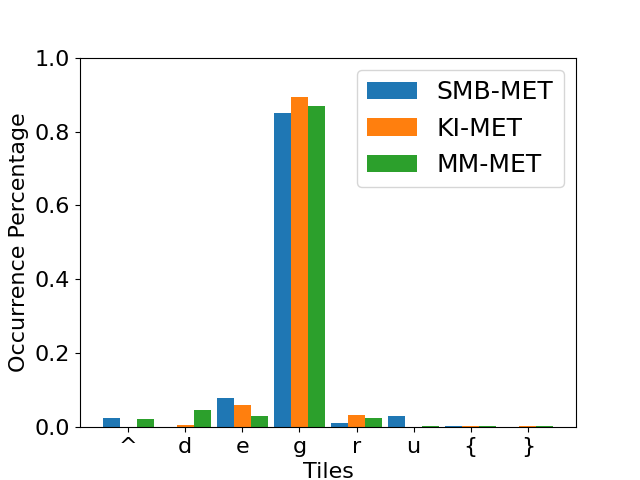}
&\includegraphics[width=0.27\textwidth]{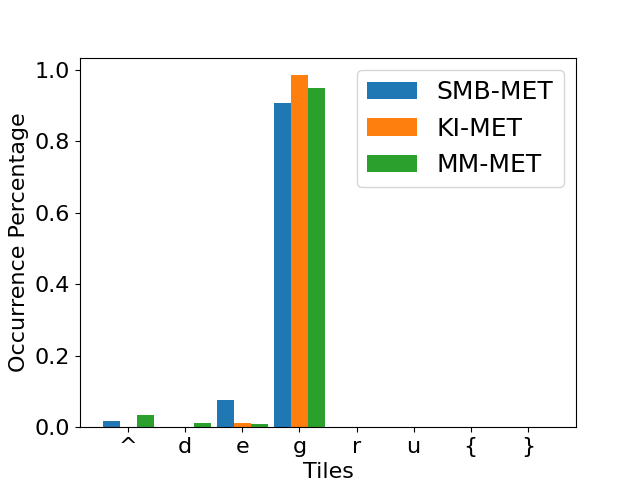}\\
Original Met & \multicolumn{2}{c}{Met as Target} \\
\end{tabular}
\caption{\label{XFIGUREtilehist} Tile histograms for each original game (left) and that game as the target using MRF-4 (center) and AE-256 (right).}
\end{figure*}
}

\newcommand{\XFIGUREaeexamplesvert}{
\begin{figure}[t!]
\centering
\setlength{\tabcolsep}{1pt}
\begin{tabular}{cc}
\includegraphics[width=0.225\textwidth]{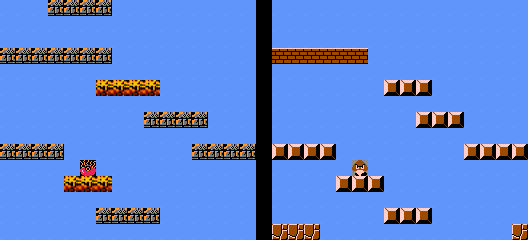}&
\includegraphics[width=0.225\textwidth]{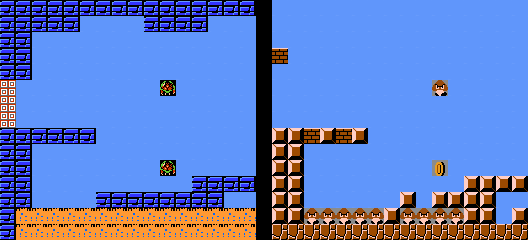}\\
KI-to-SMB & Met-to-SMB \\
\includegraphics[width=0.225\textwidth]{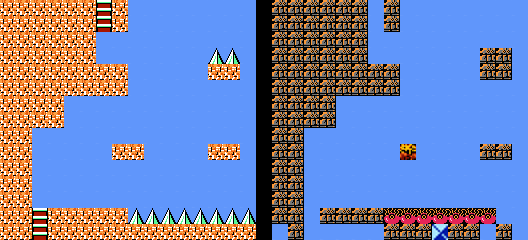}&
\includegraphics[width=0.225\textwidth]{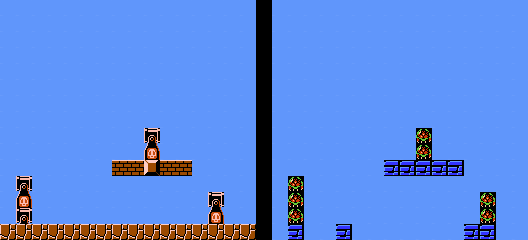}\\
MM-to-KI & SMB-to-Met\\
\includegraphics[width=0.225\textwidth]{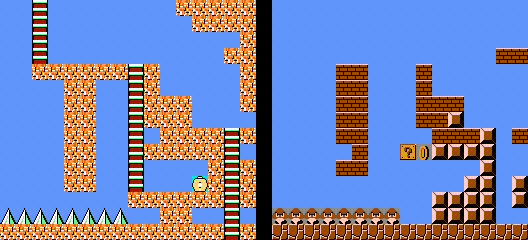}&
\includegraphics[width=0.225\textwidth]{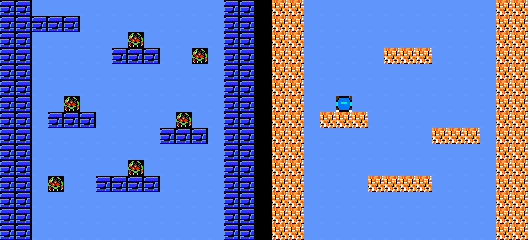}
\\
MM-to-SMB & Met-to-MM\\
\end{tabular}
\caption{\label{XFIGUREaeexamplesvert} Example style transfers using autoencoders}
\end{figure}
}

\newcommand{\XTABLEapkldiv}{\begin{table}[t!]
\small
\centering
\setlength{\tabcolsep}{1pt}
\begin{tabular}{|c|c|c|}
\hline
Source-Target     & TF-Target vs OG-Source     & TF-Target vs OG-Target   \\ 
\hline
KI-SMB & $\mathbf{0.71\pm0.61}$ & $1.52\pm1.36$ \\
MM-SMB  & $\mathbf{1.39\pm1.15}$ & $2.13\pm1.70$ \\
Met-SMB  & $\mathbf{1.32\pm1.19}$ & $2.05\pm1.63$ \\
\hline
SMB-KI & $\mathbf{0.27\pm0.22}$ & $0.87\pm0.58$ \\
MM-KI  & $\mathbf{1.34\pm1.11}$ & $1.99\pm1.6$ \\
Met-KI  & $\mathbf{1.17\pm1.02}$ & $1.72\pm1.38$ \\
\hline
SMB-MM & $\mathbf{0.45\pm0.32}$ & $1.08\pm0.62$ \\
KI-MM  & $\mathbf{0.38\pm0.36}$ & $1.8\pm0.56$ \\
Met-MM  & $\mathbf{0.82\pm0.61}$ & $1.23\pm0.98$ \\
\hline
SMB-Met & $\mathbf{0.19\pm0.16}$ & $1.22\pm0.74$ \\
KI-Met  & $\mathbf{0.38\pm0.38}$ & $2.31\pm2.14$ \\
MM-Met  & $\mathbf{0.97\pm0.74}$ & $1.67\pm1.45$ \\
\hline
\end{tabular}
\caption{\label{XTABLEapkldiv} Mean APKLDiv values along with standard deviation using AE-256 for each source-target pair. TF-Target refers to levels obtained by transferring style from original source (OG-Source) levels. Values obtained on comparing with original target (OG-Target) levels are also shown.}
\end{table}
}

\newcommand{\XTABLEplayability}{\begin{table}[t!]
\small
\centering
\setlength{\tabcolsep}{2pt}
\begin{tabular}{|c|c|c|c|c|}
\hline
& MRF-4     & MRF-8 & AE-128 & AE-256\\
\hline
KI-SMB & 75 & 67.5 & \textbf{81.25} & 75 \\
MM-SMB  & 39.86 & 43.36 & \textbf{80.42} & 77.62 \\
Met-SMB  & 33.07 & 31.65 & 72.87 & \textbf{79.54}\\
\hline
SMB-KI & \textbf{75.57} & 71.59 & 62.5 & 60.8 \\
MM-KI  & 44.76 & 46.85 & \textbf{49.65} & 48.25 \\
Met-KI  & \textbf{48.79} & 46.37 & 32.87 & 39.31\\
\hline
SMB-MM & \textbf{71.34} & 68.18 & 59.09 & 64.21 \\
KI-MM  & \textbf{69.23} & 63.75 & 60 & 67.5 \\
Met-MM  & 32.63 & 33.47 & 38.39 & \textbf{48.28}\\
\hline
SMB-Met & \textbf{85.8} & 81.25 & 60.23 & 63.07 \\
KI-Met  & 72.15 & \textbf{73.75} & 61.25 & 65 \\
MM-Met  & 46.77 & 52.45 & \textbf{61.54} & 60.14\\
\hline
\end{tabular}
\caption{\label{XTABLEplayability} Percentage of playable segments obtained for all models on each possible pair of games. Highest values per pair are in bold.}
\end{table}
}

\newcommand{\XTABLEmapping}{\begin{table}[t!]
\centering
\setlength{\tabcolsep}{1pt}
\begin{tabular}{|c|c|c|c|c|}
\hline
Aff & SMB     & KI & MM & Met\\
\hline
\texttt{X} & 
\includegraphics[width=9pt,height=9pt]{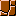} \includegraphics[width=9pt,height=9pt]{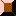} \includegraphics[width=9pt,height=9pt]{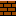} \includegraphics[width=9pt,height=9pt]{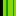} \includegraphics[width=9pt,height=9pt]{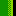} \includegraphics[width=9pt,height=9pt]{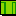} \includegraphics[width=9pt,height=9pt]{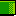}&  
\includegraphics[width=9pt,height=9pt]{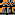}
\includegraphics[width=9pt,height=9pt]{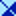}
\includegraphics[width=9pt,height=9pt]{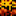}
& 
\includegraphics[width=9pt,height=9pt]{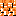}
\includegraphics[width=9pt,height=9pt]{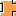}
\includegraphics[width=9pt,height=9pt]{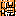}
&  
\includegraphics[width=9pt,height=9pt]{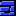}
\includegraphics[width=9pt,height=9pt]{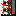}
\includegraphics[width=9pt,height=9pt]{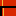}
\includegraphics[width=9pt,height=9pt]{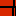}\\
\hline
\texttt{E}  & \includegraphics[width=9pt,height=9pt]{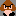} \includegraphics[width=9pt,height=9pt]{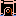} \includegraphics[width=9pt,height=9pt]{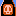} &  
\includegraphics[width=9pt,height=9pt]{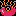}
&  
\includegraphics[width=9pt,height=9pt]{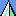}
\includegraphics[width=9pt,height=9pt]{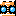}
\includegraphics[width=9pt,height=9pt]{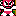}
& 
\includegraphics[width=9pt,height=9pt]{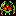}
\includegraphics[width=9pt,height=9pt]{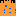} \\
\hline
\texttt{|}  & None &  \includegraphics[width=9pt,height=9pt]{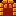} &
\includegraphics[width=9pt,height=9pt]{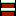}&
\includegraphics[width=9pt,height=9pt]{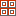} \\
\hline
\texttt{*} & \includegraphics[width=9pt,height=9pt]{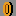} \includegraphics[width=9pt,height=9pt]{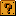} & None & 
\includegraphics[width=9pt,height=9pt]{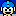}
\includegraphics[width=9pt,height=9pt]{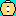}
\includegraphics[width=9pt,height=9pt]{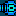}
\includegraphics[width=9pt,height=9pt]{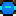}
\includegraphics[width=9pt,height=9pt]{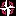}
&
None
\\
\hline
\end{tabular}
\caption{\label{XTABLEmapping} Tile-to-affordance mapping. Background (\texttt{-}) was the same for each game.}
\end{table}
}


\section{Introduction}
Machine learning-based methods have come to be increasingly applied for creative applications, particularly in the domain of visual art. One such popular application has been neural style transfer---the use of convolutional neural nets for rendering the content within an image in various styles. Since being introduced by \citeauthoryearp{gatys2016neural}, a large body of research has centered around improving and experimenting with various methods for style transfer \cite{jing2019neural} as well as developing user-facing style transfer tools \cite{champandard2016semantic}. Models such as CycleGAN \cite{zhu2017unpaired} and pix2pix \cite{isola2017image} have become especially popular for their ability to learn translation functions between sets of images, thereby enabling a wide variety of artistic applications and installations.

Relatedly, a subset of Procedural Content Generation via Machine Learning (PCGML) \cite{summerville2018procedural} methods have come to focus on more creative applications of ML beyond just generating levels for a given game. These have centered around reasoning about the design spaces of multiple games taken together and have involved domain transfer \cite{snodgrass2016approach} and combinational creativity-based methods for content generation \cite{guzdial2018combinatorial,sarkar2020exploring}. \citeauthoryearp{sarkar2020towards} recently proposed the term Game Design via Creative Machine Learning (GDCML) to classify this creative subset of PCGML, further calling for taking inspiration from existing creative ML approaches in visual art and music to inform new creative ML approaches for games. One such application is style transfer for games. Despite the recent rise of GDCML works, videogame style transfer has remained underexplored. Traditional style transfer methods in the visual art and image domains are pixel-based and thus not well suited for games. In having to satisfy functional constraints, games in several PCGML works have typically benefited from precise, discretized tile-based representations. Thus, a style transfer method suited for games could also utilize such tiled representations.

In this work, we introduce \textit{tile2tile}, an approach for style transfer between tile-based platformer games. Our method is inspired by pix2pix but instead of learning to translate between sets of images, we learn to translate between sets of tile-based levels, hence the name. More specifically, we train models to convert levels from a game-agnostic form based on tile affordances to game-specific tile representations. Then by converting a level from game A to the sketch form and then using the model to convert the resulting sketch form to a level from game B, we obtain a method for transferring style between games. We demonstrate our approach using Markov random fields and convolutional autoencoders to transfer style between levels of the platformer games \textit{Super Mario Bros.}, \textit{Metroid}, \textit{Mega Man} and \textit{Kid Icarus}. This work thus contributes:
\begin{enumerate}
    \item An approach for style transfer between tile-based games
    \item A demonstration of the approach using Markov random fields and convolutional autoencoders
\end{enumerate}

\XFIGUREtiletile

\section{Background}

Neural style transfer was introduced by the seminal work of \citeauthoryearp{gatys2016neural} and has since been the subject of a wide variety of research, with a survey given by \citeauthoryearp{jing2019neural}. While traditionally this referred to transferring styles of individual images, the pix2pix model \cite{isola2017image} enabled the notion of learning translation functions between sets of images by training on pairs of images, in turn allowing for style transfer-adjacent applications between different image domains. We use a similar paired approach for training our models, but operate in the tile domain of game levels rather than the pixel domain of images. To our knowledge, the only prior work attempting style transfer within games is the work of \citeauthoryearp{dadfar2021gan} and a demo by Google Stadia \cite{stadia_demo} but both focus only on transferring visual image style on to the game without considering affordances and are pixel-based rather than tile-based.

Related to style transfer, a key focus of recent creative PCGML research has been to learn, reason about and leverage design spaces spanning a number of games, rather than learn distributions of individual games. This has involved generating new types of content by recombining learned game graphs \cite{guzdial2018automated}, blending latent representations \cite{sarkar2019blending}, learning affordance-based tile embeddings across multiple games \cite{jadhav2021tile} and performing domain adaptation \cite{snodgrass2016approach}. The last of these is particularly related to our work but differs in learning direct mappings between the levels of two games and focusing on generating new levels for the target domain, whereas we learn models that map a unified affordance-based representation to the tiles specific to a game and focus on transferring the style of a given level. Additionally, we refer to levels in the affordance-based representation as sketches, borrowing vocabulary from \cite{snodgrass2019levels, snodgrass2020multi} who use sketches to refer to lower-resolution level representations indicating only solid and non-solid tiles. The set of affordances we use is a simplified subset of those used by \citeauthoryearp{sarkar2020exploring} to blend levels of multiple platformers. Their tile-based affordances were in turn inspired by the pixel-based affordance work of \citeauthoryearp{bentley2019videogame}. Recent game blending work \cite{sarkar2020exploring,sarkar2020conditional} i.e. generating new games by blending levels/mechanics of existing ones, is related to style transfer but differs in that it focuses on blending together different games where as we learn to map from one game to another. Closely related to our work is also the recent work of \citeauthoryearp{chen2020image} who generated Mario levels using an input image, thus performing a version of style transfer. Our work differs in attempting style transfer between game levels. Similar to them, we also use Markov models and autoencoders, both of which have been used in several prior PCGML works. \citeauthoryearp{snodgrass2017learning} used both Markov chains and Markov random fields for generating platformer levels while autoencoders have been employed for level repair \cite{jain_autoencoders_2016} as well as learning level design patterns \cite{guzdial2018explainable}.

\section{Method}

\subsubsection{Overview}
The overall tile2tile approach is depicted in Figure \ref{XFIGUREtiletile}. This method of transferring style between levels of two games A (source) and B (target) consists of:
\begin{enumerate}
\item Converting the source level from A into a low-resolution representation \textit{(sketch)} based on tile affordances
\item Translating the level sketch into the original tile representation of game B using a trained model \textit{(filter)}
\end{enumerate}

Thus, central to the approach is the concept of tile affordances. In general, affordances \cite{bentley2019videogame} for an object refer to the types of in-game actions and interactions that it permits. The key idea that enables our proposed approach is that while games differ in their specific tile-based representations, they can be considered to share a unified affordance-based representation. For example, goombas appear in Mario while metroid creatures appear in Metroid but both of these are enemies that can harm the player; similarly, Mario features coins while Metroid features various powerups but both can be considered as having the affordance of being collectible. Hence by defining a common affordance-based level representation shared by a set of games, and combining it with a model that maps from affordances to the specific tiles of a game, we obtain a method of style transfer by converting a level from a source game to the affordance representation and applying the model for the target game to map this affordance representation to the original tiles of the target game. 

We can view this notion of affordances and tiles as analogous to the classic style transfer concepts of \textit{content} and \textit{style} where content refers to the underlying structures and objects within an image while style refers to the high-level visual characteristics such as color, paint style, textures etc. Similarly, we can view affordances as representing the underlying game-agnostic structures and topology such as generic solid tiles and collectibles (i.e. the content) and the tiles as representing the high-level game-specific attributes such as pipes in Mario and powerups in Metroid (i.e. the style). Thus, under this conception of style transfer, when transferring style from one game to another, our goal is to keep the underlying affordances similar while transforming the high-level game-specific tiles.

To perform this, for each game, we define a fixed mapping that translates each of its original tiles to a tile in the common affordance representation, shared across all games. By applying this mapping, we obtain lower-resolution versions of the original levels for each game, which we refer to as \textit{sketches}. Then, for each game, we train a model to convert the game's sketches back to its original tile representation. We refer to these models as \textit{filters} for their respective games, analogous to image filters which similarly transform the visual characteristics of images. Since each game's filter converts a sketch into that game's tile representation, we hence obtain a style transfer process by converting a level from a source game into a sketch and then applying the target game's filter on that sketch to obtain the same source level but in the target game's tile representation. In this work, we implement filters using Markov random fields and convolutional autoencoders.

\XFIGUREtileaff

\XTABLEmapping

\subsubsection{Tile and Affordance Data}
We used levels of 4 platformer games---\textit{Super Mario Bros (SMB)}, \textit{Kid Icarus (KI)}, \textit{Mega Man (MM)} and \textit{Metroid (Met)}---all taken from the Video Game Level Corpus (VGLC) \cite{summerville2016vglc}. VGLC levels are in text format with different characters representing different tiles. Since there is overlap in terms of the same characters being used to represent different tiles across games, we edited the original tilesets to ensure that each game had a completely distinct set of tiles. For the affordance tileset common to all games, we opted to use a much simplified version of the set of affordances used by \citeauthoryearp{sarkar2020exploring}, settling on the following:
\begin{itemize}
    \item \texttt{X}: \textit{solid} (e.g., ground, platforms, breakable)
    \item \texttt{|}: \textit{passable, climbable}, (e.g., doors, ladders)
    \item \texttt{E}: \textit{hazard, enemy}, (e.g., goombas, spikes, lava, metroids)
    \item \texttt{*}: \textit{collectable}, (e.g., coins, powerups)
    \item \texttt{-}: \textit{empty}, (e.g. background)
\end{itemize}

Example levels in their original and affordance representations are shown in Figure \ref{XFIGUREtileaff}. For each game, we defined a hand-authored mapping from each original tile in the game to one of the above affordances.
The mapping is shown in Table \ref{XTABLEmapping}. The number of distinct tiles in SMB, KI, MM and Met were 14, 6, 16 and 8 respectively. Since the number of distinct affordance tiles in this work is only 5, the tile-to-affordance mapping is many-to-one and hence this translation can be done manually, as the conversion is deterministic. However, each affordance tile could map to several game tiles and thus we need a model that can appropriately translate affordance sketches to levels for a particular game. For this, we test two approaches---Markov random fields (MRFs) and convolutional autoencoders. The MRFs were trained directly using whole levels from the VGLC. Autoencoders however work with fixed-size inputs and outputs. Since VGLC levels vary in size and are limited in number, we trained the autoencoders on segments of levels, as is typical in most neural net-based PCGML work. For this, we extracted 15x16 segments from each of the 4 games. This was determined based on MM and Met having levels whose horizontal and vertical sections are 15 tiles high and 16 tiles wide respectively. KI levels are also 16 tiles wide while SMB levels are originally 14 tiles high so we added a row of background tiles as padding at the top of each level. Each tile in a segment was converted to a one-hot encoding. Thus, the input dimensions for a segment were nx15x16 where n is the number of unique tiles in that game. To extract segments, we slid a 15x16 window across each level, one row and column at a time, giving us 2643, 1171, 3118 and 3762 segments for SMB, KI, MM and Met respectively. For Met, we extracted segments only from level 3 of the VGLC.

\XFIGUREmrf

\XFIGUREae

\subsubsection{Markov Random Field}
Our first approach to training filters involves using markov random fields (MRFs) \cite{clifford1990markov}. While markov chains (MCs) learn conditional probability distributions (CPDs) for the current state given one or more preceding states, MRFs learn CPDs for the current state given its surrounding states. Though both MCs and MRFs have been used previously for level generation \cite{snodgrass2017learning}, we opt to use MRFs since for the task of style transfer, we already have a source level to begin with and thus have a set of surrounding tiles for each position, unlike when generating levels from scratch where typically we would only have preceding tiles. Using an MRF allows us to consider a given tile's dependencies in all directions. Note that in the proposed tile2tile framework, the role of the model for a given game is to learn to map affordance tiles to tiles specific to that game i.e. convert a sketch to a level for that game.
Using an MRF, we model this as learning the probability of a location in the target level being a certain tile given the affordances of the surrounding neighborhood of that same location in the source level. We refer to this neighborhood as the context for the specific central tile. 
Thus, the MRF for a game learns the probability of tile occurrences given a surrounding affordance context. We test two contexts - one looking at only the 4 affordance tiles north, south, east and west of the central tile (MRF-4) and the other looking at the full 8-tile surrounding context (MRF-8). An example is shown in Figure \ref{XFIGUREmrf}.

Training an MRF model consists of two steps---1) determining absolute counts and 2) estimating probability distributions. We implement MRFs as dictionaries with each unique context serving as a key and each key mapping to the distribution of center tiles for that context. To train MRFs for a given game, we first convert each level to its sketch form using that game's pre-defined tile-to-affordance mapping. Then, we slide a 3x3 window across each level, moving one column horizontally and/or one row vertically at a time, updating the counts of original center tiles (i.e. before converting to sketch form) for each context that we find along the way. Thus, after this step, the dictionary stores the number of times each original game tile appears in the center for each unique sketch context. In the final step, we  convert each context's absolute counts into a distribution that we can sample from when performing style transfer. After training, we can use the trained MRF for a game T to change the style of levels from other games to that of T using the pipeline shown in Figure \ref{XFIGUREtiletile}. That is, we convert the source level into sketch form using the tile-to-affordance mapping and then similar to training, slide a 3x3 window across the level and replace the center tile by sampling from the trained MRF. More specifically, we look up the MRF's dictionary using the context and then sample a tile according to the learned distribution. If the context is not found (i.e. was not seen during training), we randomly sample a tile with the same affordance as the current center tile in the source level. Example levels are shown in Figures \ref{XFIGUREsmbtoki} and \ref{XFIGUREmmtomet}. Note that for MRFs, the style transfer process by design explicitly keeps the affordances fixed from source to target.

\subsubsection{Autoencoder}
Our second approach to learning filters is through the use of convolutional autoencoders. 
Autoencoders \cite{hinton2006reducing} consist of encoder and decoder neural networks. The encoder maps the input data to a compressed, hidden representation which is then used by the decoder to learn to reconstruct the original input in an unsupervised manner. Convolutional autoencoders feature encoders and decoders made up of convolutional rather than fully-connected layers.
For training the autoencoders for each game, we collected pairs of level segments from that game and their corresponding sketch representations. We then used the sketches as the encoder inputs but computed the loss on the decoder outputs using the original segment. Thus, through training, the model learned to convert a sketch of a segment to its original tile representation. This training process is depicted in Figure \ref{XFIGUREae}. 
Our approach is analogous to denoising autoencoders \cite{vincent2008extracting} where the encoder input is a noisy form of the original and loss is computed between the decoder output and the original input. We use a translated input where instead of adding noise, we translate to affordances. The underlying structures in inputs/target outputs are identical with only the representation differing.
After training, the autoencoders can perform style transfer similarly to MRFs. We take a 15x16 segment of the source game, convert it into sketch form, forward it through the layers of the autoencoder for the target game which finally outputs the segment translated to the tiles of the target game. Example levels are shown in Figure \ref{XFIGUREaeexamplesvert}. In contrast to the MRF, the process does not keep the affordances fixed explicitly since the autoencoder works at the segment-level rather than at the tile-level.
We use a convolutional autoencoder implemented using Pytorch \cite{paszke2017automatic}. The encoder and decoder consisted of 3 convolutional and 3 transpose convolutional layers respectively, with both using Batch-Normalization and ReLU activation. The encoder inputs were of dimension 5x15x16(=1200) with the 5 channels corresponding to there being 5 unique affordance tiles. The decoder outputs were of dimension (nx15x16) where n varied based on the number of unique tiles in the particular game. Using the format (in\_channelsxout\_channels, kernel\_size, stride\_length), the encoder layers were (1200x512, 4, 1), (512x256, 4, 1) and (256xhidden\_size, 4, 2) while the decoder layers were (hidden\_sizex64, 4, 2), (64x128, 4, 1), (128x(nx15x16), 4, 1). Models were trained for 250 epochs using binary cross entropy loss and the Adam optimizer with an initial learning rate of 0.001, decayed by 0.1 each time the training loss hit a plateau for 50 epochs. We experimented with several hidden sizes finding 128 and 256 dimensions to most consistently capture the affordance structures of the source game. For the rest of the paper, we thus use the 256 (AE-256) and 128-dimensional (AE-128) versions.

\section{Evaluation}
Due to the subjective nature of style and aesthetics, evaluating style transfer is a challenging problem \cite{jing2019neural}. Ideally we could perform a qualitative user study and elicit opinions from observers regarding the goodness of style transfer but for this initial work, we perform a quantitative evaluation focusing on affordances and tiles.

\XTABLEapkldiv

\XFIGUREtilehist

\XFIGUREaeexamplesvert

\subsubsection{Content} As mentioned before, style transfer is often described in terms of \textit{content} and \textit{style}. Typically, the goal is to maintain the content of the source image while applying the desired new style in the target image. This distance between the content of the source and target image is the content loss. Analogously, when performing style transfer on a level from a source to a target game, we want the underlying affordances related to the overall level topology and structure (i.e. content) to be close to the source while changing the higher-level tile representations (i.e. style) to that of the target. Thus, we compare the tile affordance patterns in the input source level and the output target level. For this, we use the Tile Pattern KL-Divergence metric \cite{lucas2019tile} but use the affordance tiles, not the original game tiles. To avoid confusion, we refer to this as the Affordance Pattern KL-Divergence metric (APKLDiv) in this work. Thus, APKLDiv can be viewed as being analogous to the content loss from visual art style transfer. For MRFs, since the style transfer process by design keeps the affordances fixed from source to target, this evaluation would not be useful. However, it is a useful evaluation for autoencoders since they have to learn to keep the affordances the same.

To perform this evaluation, for every pair of games S (source) and T (target), we apply T's autoencoder on all original segments of S. This gives us 2 sets of segments---the original segments of S and the generated segments of the target game T.
We then compute the APKLDiv values between these 2 sets using their sketch forms, averaging values obtained for patterns of size 2x2, 3x3 and 4x4. The method for computing these values is given in \cite{lucas2019tile}. An additional, useful set of levels to consider is the original levels for the target game T, also in sketch form. Consider the example of an SMB-to-KI conversion where SMB is the source S and KI is the target T. For effective style transfer, we want the APKLDiv values between the style-transferred KI segments and the original SMB segments to be lower since the goal is to maintain the affordances of the original SMB segments while transferring style to KI. On the other hand, the distance between the style-transferred KI segments and the original KI segments should be higher because the affordances of the transferred levels should be closer to the source SMB segments than the original KI segments. This shows that the autoencoder is preserving the content/affordances of the source level rather than just generating the levels it was trained on. Note that each autoencoder was only trained on levels for its target game.
Results for AE-256 are shown in Table \ref{XTABLEapkldiv}. In all cases, the distance between the style-transferred levels and the original source levels is lower than the distance between the style-transferred levels and the original target levels, as we would expect. For space, we omit results for AE-128 but they followed the same patterns except values in both columns were higher (i.e. it did slightly worse than AE-256). Overall, this evaluation shows that the autoencoder for the target game is able to preserve the content of levels from the source game when tasked with performing style transfer.
As a concrete example, consider the Met-to-SMB example in Figure \ref{XFIGUREaeexamplesvert}. The underlying affordances of the SMB level are uncharacteristic of SMB and are more representative of Metroid. Thus, such SMB levels translated from Metroid would be closer to original Metroid than original SMB levels in terms of their affordances.

\subsubsection{Style} 
If we consider the underlying affordance-based level patterns and structures to capture the game-agnostic content of a level, then the game-specific style is defined by the game-specific tile representation of those affordance patterns. For example, consider a level from a source game S which is style-transferred to target game T. If we compare this new game T level with an original level from game T, we expect their underlying content (i.e. affordance patterns) to be different since the former was converted from game S. For the style to be consistent, we want the tile distributions to be similar, since they both now represent game T. To evaluate this, we consider the game-specific tile distributions and compare tile histograms between the generated levels obtained via style transfer and the original levels of that game. For each game, we computed the tile histogram (i.e. the discrete distribution of tiles) using all the original levels in the game. Then for each pair of games (S, T) where S is the source and T is the target, we applied the model for T on each level of S and computed the tile histogram using all style-transferred levels. For space we show results for MRF-4 and AE-256 in Figure \ref{XFIGUREtilehist}. While not perfect, the tile distributions obtained after style transfer are similar to those of the original levels. We do not expect these to be identical since some amount of difference is a consequence of trying to match the affordances of the source. The biggest difference is observed for the ground G tile in SMB. This may be due to SMB being the only one with separate solid tiles for the ground (G). Additionally, due to having vertical sections, the other 3 games have several segments without a ground row. Hence it is reasonable that when converting from such segments, the resulting SMB segments have lower proportions of ground tiles than the originals.

\XFIGUREsmbtoki

\XFIGUREmmtomet

\subsubsection{Playability}
Finally, we evaluated the playability of style-transferred levels. Though MRFs can work with whole levels, since autoencoders can only work with fixed-size segments, for an even comparison, we evaluated the MRF playability with segments as well. Using whole levels is additionally complicated by the A* agents we use for determining playability requiring specific start and goal tiles to be identified, which is not ideal for a Metroid map, due to backtracking. Thus, we evaluated both models using 15x16 segments on which we ran the game-specific tile-based A* agents from \cite{sarkar2020exploring} where each game's agent is capable of performing the specific jumps in that game. For each source-target game pair, we applied the model for the target game on every level of the source game and then determined playability by running the target game's agent. A segment is playable if the agent is able to find a path from start to goal. For each segment, we ran the agent in both the horizontal and vertical direction. A segment is playable if either a horizontal or vertical path is found. 
For each model and source-target game pair, we computed the percentage of source levels that were playable after style transfer. Results are shown in Table \ref{XTABLEplayability}. We see that performance varies for different game pairs. Autoencoders do best when the target is SMB as well as when the source is MM. MRFs do better when SMB is the source but do poorly when Met is involved in the transfer though it does well for Met-KI. While future work analyzing playability after style transfer would be useful, it is reasonable that transferring style would lead to a dip in playability and we still obtain reasonably high playability percentages for several pairs. In the future, we could incorporate repair capabilities into the playability agents such as in \cite{cooper2020pathfinding} as well as explore developing generalized agents capable of playing both source and target levels which could in turn lead to generating new types of mechanics.

\XTABLEplayability

\section{Conclusion and Future Work}
We introduced tile2tile, a style transfer approach for tile-based platformer levels, and demonstrated it using Markov random fields and convolutional autoencoders. In the future, we wish to further validate this approach using user studies and develop an application for users to apply the trained filters to perform style transfer on pre-existing or hand-authored levels. Additionally, we could apply the approach to augment the VGLC by adding style-transferred levels to the corpus and in general use style-transferred levels to supplement levels for games with insufficient training data. A further direction would be to generalize the method to other genres. Our approach should work for tile-based games broadly as long as a common set of affordances can be defined between source and target games e.g. a tile-to-affordance mapping for dungeon-based levels could be used to train a model for translating Zelda levels to maps for some other tile-based dungeon crawler. 

\bibliography{refs-custom}

\end{document}